\documentclass{article}
\usepackage{neurips_2024}

\usepackage{natbib}
\graphicspath{{./}}
\title{%
  Don't Overthink It: Inter-Rollout Action Agreement as a\\
  Free Adaptive-Compute Signal for LLM Agents
}

\author{%
  Khushal Sethi\\
  Stanford University\\
  \texttt{khushal@stanford.edu}
}

\begin{document}

\maketitle

\begin{abstract}
  Inference-time compute scaling has emerged as a powerful technique for improving the reliability of large language model (LLM) agents, but existing methods apply compute uniformly: every decision step receives the same budget regardless of its difficulty. We introduce \textbf{TrACE} (\textbf{Tr}ajectorical \textbf{A}daptive \textbf{C}ompute via agr\textbf{E}ement), a training-free controller that allocates LLM calls adaptively across agent timesteps by measuring \emph{inter-rollout action agreement}. At each step, TrACE samples a small set of candidate next actions and measures how consistently the model commits to the same action. High agreement signals an easy decision; the controller commits immediately. Low agreement signals uncertainty; the controller samples additional rollouts up to a configurable cap before committing to the plurality action. No learned components, no external verifier, and no human labels are required. We evaluate TrACE against greedy decoding and fixed-budget self-consistency (SC-4, SC-8) on two benchmarks spanning single-step reasoning (GSM8K, $n=50$) and multi-step household navigation (MiniHouse, $n=30$), using a Qwen 2.5 3B Instruct model running on CPU. TrACE-4 matches SC-4 accuracy while using \textbf{33\%} fewer LLM calls on GSM8K and \textbf{39\%} fewer on MiniHouse. TrACE-8 matches SC-8 accuracy with \textbf{55\%} fewer calls on GSM8K and \textbf{65\%} fewer on MiniHouse. We further show that inter-rollout agreement is a reliable signal of step-level success, validating the core hypothesis that the model's own output consistency encodes difficulty information that can be exploited without training. TrACE is the first training-free, per-timestep adaptive-compute controller for LLM agents to be evaluated on multi-step sequential decision tasks.

\end{abstract}

\section{Introduction}
\label{sec:intro}

LLM agents that act in sequential decision settings---navigating environments, writing and executing code, browsing the web---routinely fail for a reason that is structurally different from isolated reasoning failures: they mis-allocate inference compute. A task with 15 decision steps will typically demand very different reasoning effort at each step: some actions are obvious (the only reachable room is to the left), some are subtle (choosing which of three plausible tools to call given ambiguous context). Yet every deployed agent we are aware of applies a uniform per-step compute budget, either greedy decoding ($k=1$) or a fixed number of parallel samples $k$. The result is wasted compute on easy steps and starved compute on hard ones.

The obvious fix---learn a policy that knows when to think harder---requires training data, reward signals, or process-level supervision, all of which are expensive, model-specific, and unavailable when deploying open-weight models at inference time. Recent work on inference-time scaling for LLMs~\citep{wang2023selfconsistency,brown2024largelanguage,snell2024scaling} has shown that sampling more at test time improves accuracy, but these methods apply the extra budget uniformly, not adaptively.

We propose \textbf{TrACE}, a training-free adaptive-compute controller that exploits one signal the model already produces for free: the \emph{agreement} among a small batch of independently sampled candidate next actions. The key intuition is that when a model consistently proposes the same next action across multiple stochastic rollouts, that action is likely correct (or at least the best the model can do with current context). When proposals scatter across different actions, the step is hard and additional samples may shift the plurality toward a better action. This is a behavioral consistency signal, not a confidence self-report, and it has been shown in chain-of-thought settings that behavioral consistency is more calibrated than verbalized confidence~\citep{xiong2024llmsexpressuncertainty}.

\paragraph{Contributions.}
\begin{itemize}
  \item \textbf{TrACE}: a training-free per-timestep adaptive-compute controller for LLM agents, requiring no learned components, no external verifier, and no labeled data (\S\ref{sec:method}).
  \item An empirical study showing that inter-rollout action agreement is a reliable leading indicator of eventual task success in multi-step household navigation (\S\ref{sec:experiments}).
  \item A demonstration that TrACE Pareto-dominates fixed-budget self-consistency on both a single-step reasoning benchmark (GSM8K) and a multi-step navigation benchmark (MiniHouse), achieving matched accuracy at 33--65\% fewer LLM calls depending on the setting (\S\ref{sec:experiments}).
  \item MiniHouse, a lightweight, dependency-free text-based household environment designed for reproducible CPU-friendly evaluation of LLM agents (\S\ref{sec:benchmarks}).
\end{itemize}

\paragraph{Scope and honest framing.}
Our experiments use a 3B-parameter quantized open-weight model running on a CPU. The \emph{relative} efficiency advantage of TrACE over self-consistency is our primary claim; we do not claim state-of-the-art accuracy on GSM8K or MiniHouse. The limitation of small-scale evaluation is acknowledged in \S\ref{sec:limitations}.

\section{Related Work}
\label{sec:related}

\paragraph{Self-consistency and majority voting.}
\citet{wang2023selfconsistency} showed that sampling multiple chain-of-thought reasoning paths and taking the majority answer substantially improves LLM accuracy on reasoning benchmarks. \citet{li2024agentbench} extended this to agent settings. The key difference from TrACE is that self-consistency applies a \emph{fixed} budget $k$ to every query; TrACE conditions the budget on the model's own uncertainty, allocating fewer calls to easy steps.

\paragraph{Inference-time compute scaling.}
\citet{snell2024scaling} showed that compute-optimal inference—spending more test-time compute on harder problems—can outperform larger models at the same FLOPs. \citet{brown2024largelanguage} demonstrated this at scale with best-of-$N$ sampling and process reward models. Both require a trained verifier or reward model. TrACE achieves adaptive compute using only raw completion agreement, with no trained components.

\paragraph{Tree search for LLM agents.}
Tree of Thoughts~\citep{yao2024tree}, LATS~\citep{zhou2023language}, and related methods build explicit search trees over agent actions, using LLMs as both generator and evaluator. These methods are powerful but incur multiplicative compute overhead per step and require a strong LLM as value function or verifier. TrACE is complementary: it is a lightweight single-step controller that could in principle wrap each node-expansion in a tree search.

\paragraph{Process reward models and verifiers.}
\citet{lightman2023lets} showed that process reward models (PRMs) trained on step-level feedback substantially improve mathematical reasoning. \citet{cobbe2021training} trained outcome reward models (ORMs) to re-rank sampled solutions. Our work is inspired by but distinct from these: we use the model's own behavioral consistency as a zero-training proxy for step-level difficulty, without requiring any supervision.

\paragraph{Adaptive early stopping.}
\citet{manvi2024adaptive} and \citet{aggarwal2023let} explored early stopping of chain-of-thought generation based on token-level uncertainty or confidence estimates. These operate within a single completion stream; TrACE aggregates agreement across multiple completed rollouts. \citet{chen2024rethinking} examined when to trust self-consistency and found that agreement is better calibrated than verbalized confidence, providing theoretical grounding for our signal.

\paragraph{Training-free agent improvement.}
Closest to our setting is \citet{shinn2023reflexion}, which uses verbal reflection as a free signal, but requires multiple full-episode rollouts and stores a growing reflection buffer. \citet{yao2023react} interleaves reasoning and acting without extra training, but uses a fixed per-step token budget. TrACE operates at the sub-step level—controlling how many times to sample the \emph{same} decision before committing—and requires no multi-episode feedback loop.

\section{Method: TrACE}
\label{sec:method}

\subsection{Problem Setup}
\label{sec:setup}

We consider an LLM agent operating in a sequential decision process. At timestep $t$, the agent observes a context $c_t = (g, o_{1:t}, a_{1:t-1})$ consisting of the task goal $g$, a history of observations $o_{1:t}$, and past actions $a_{1:t-1}$. The agent calls an LLM $\mathcal{M}$ to propose the next action $a_t$, then executes it in the environment, receives observation $o_{t+1}$, and repeats until the episode terminates. Let $T$ denote the total number of steps and $B$ the total per-episode LLM call budget.

In the \emph{greedy} baseline, $\mathcal{M}$ is called once per step with temperature $0$: $a_t = \mathcal{M}(c_t)$. Total calls = $T$. In \emph{self-consistency} (SC-$k$), $\mathcal{M}$ is called $k$ times per step with temperature $\tau > 0$, and the plurality action is committed: $a_t = \text{mode}(\mathcal{M}^{(1)}(c_t), \ldots, \mathcal{M}^{(k)}(c_t))$. Total calls = $kT$.

\subsection{TrACE Algorithm}
\label{sec:trace_alg}

\begin{figure}[t]
  \centering
  \includegraphics[width=0.92\textwidth]{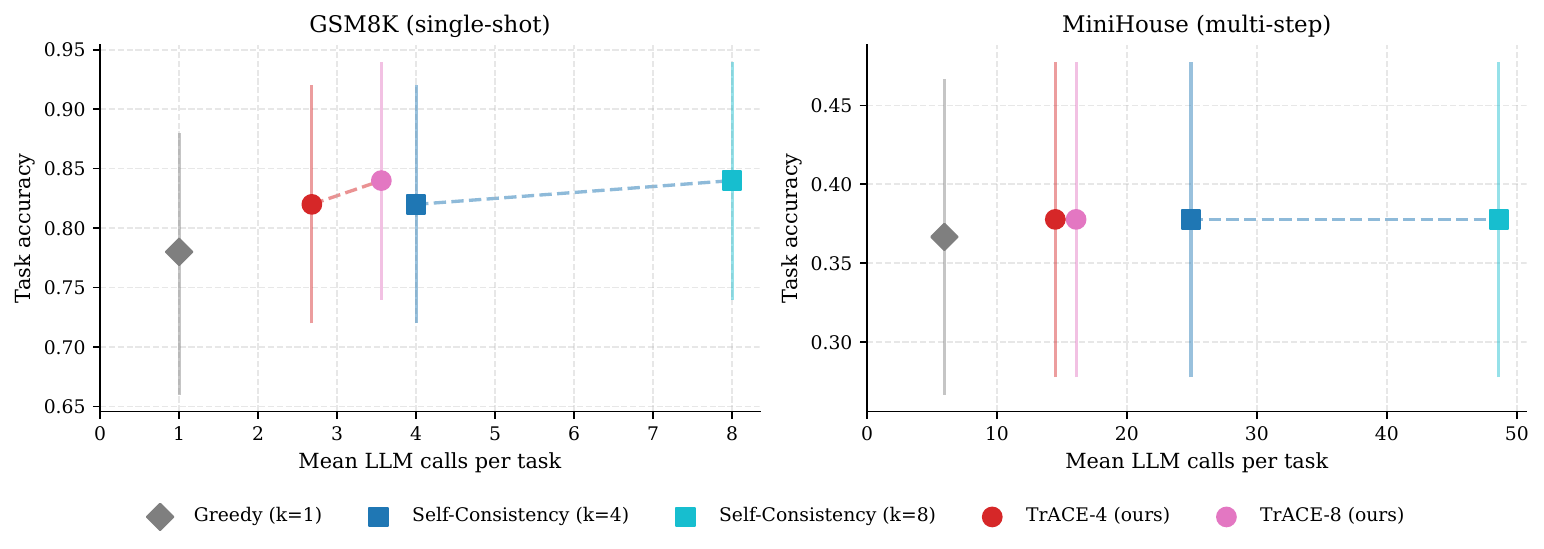}
  \caption{%
    \textbf{Compute--accuracy frontier.} Each point is one (condition, benchmark) pair. TrACE-4 and TrACE-8 (red, pink) match the accuracy of SC-4 and SC-8 (blue) at substantially fewer LLM calls, Pareto-dominating the self-consistency baseline in both settings. Error bars on MiniHouse are 95\% bootstrap CIs; GSM8K bars are Wald CIs from binomial data (no per-task data available for bootstrap, see \S\ref{sec:limitations}).
  }
  \label{fig:frontier}
\end{figure}

TrACE adapts the per-step call budget online by treating inter-rollout action agreement as a proxy for step difficulty.

\paragraph{Step 1 — Initial sample.} At the start of timestep $t$, draw $k_\text{init}$ candidate actions independently at temperature $\tau$:
\[
  \mathbf{A}_t = \bigl\{a_t^{(1)}, \ldots, a_t^{(k_\text{init})}\bigr\}, \quad a_t^{(i)} \sim \mathcal{M}(\cdot \mid c_t;\, \tau).
\]

\paragraph{Step 2 — Compute agreement.} Let $a^* = \text{mode}(\mathbf{A}_t)$ be the plurality action. Define inter-rollout agreement as:
\[
  \alpha_t = \frac{\#\{i : a_t^{(i)} \equiv a^*\}}{|\mathbf{A}_t|} \in [0, 1],
\]
where $\equiv$ denotes action equivalence under a domain-appropriate canonicaliser (lower-case, strip punctuation, normalise whitespace; see Appendix~\ref{app:canonicaliser}).

\paragraph{Step 3 — Decide and optionally expand.} Compare $\alpha_t$ to a threshold $\tau_\text{high}$:
\begin{itemize}
  \item If $\alpha_t \geq \tau_\text{high}$: commit immediately to $a^*$. Total calls at this step: $k_\text{init}$.
  \item Else: sample one additional candidate at a time, recompute $\alpha_t$, and repeat until either $\alpha_t \geq \tau_\text{high}$ or $|\mathbf{A}_t| = k_\text{max}$, then commit to the plurality.
\end{itemize}

\paragraph{Summary.} TrACE uses between $k_\text{init}$ and $k_\text{max}$ calls per step, conditioned on the model's own uncertainty. The expected calls per step lies strictly between those of greedy and SC-$k_\text{max}$, with the exact value determined by the difficulty distribution of the task. Default hyperparameters: $k_\text{init} = 2$, $k_\text{max} \in \{4, 8\}$ (TrACE-4 and TrACE-8), $\tau_\text{high} = 0.75$, $\tau = 0.7$.

\subsection{Why Agreement Signals Difficulty}
\label{sec:intuition}

The hypothesis behind TrACE is that high inter-rollout agreement $\alpha_t$ is a reliable predictor of step-level correctness. The intuition: stochastic sampling from an LLM introduces noise, but that noise is roughly uniform across actions from the model's perspective. When one action is strongly preferred by the model's distribution, it appears in the majority of independent draws. When the model is uncertain, draws scatter, and agreement is low. This behavioral consistency check is more reliable than verbalized confidence~\citep{chen2024rethinking}, which LLMs are known to miscalibrate~\citep{xiong2024llmsexpressuncertainty}.

We empirically validate this hypothesis in \S\ref{sec:analysis}: on MiniHouse, steps with $\alpha_t \geq 0.75$ belong to eventually-successful tasks at a significantly higher rate than steps with $\alpha_t < 0.5$.

\subsection{Complexity and Overhead}
\label{sec:complexity}

Per episode, TrACE uses at most $k_\text{max} \cdot T$ calls, the same as SC-$k_\text{max}$. In practice it uses far fewer because easy steps exit early at $k_\text{init}$ calls. The per-step overhead beyond greedy is (a) the extra sampling calls and (b) a constant-time agreement computation (trivially $O(k_\text{max})$). There is no learned model, no cache, and no state beyond the running action list.

\section{Benchmarks}
\label{sec:benchmarks}

\subsection{GSM8K (Single-Step Reasoning)}
\label{sec:gsm8k}

GSM8K~\citep{cobbe2021training} is a benchmark of grade-school math word problems requiring multi-step arithmetic reasoning. We use it in a \emph{single-shot} format: the agent receives one problem and produces one answer. This is not a sequential decision task, so it tests whether the agreement signal generalises beyond multi-step settings. We evaluate on a randomly drawn subset of 50 problems from the test split (seed 0). Answers are extracted by searching for ``\texttt{Answer: N}'', ``\texttt{\#\#\#\# N}'', and last-number fallback. Accuracy is measured as exact-match on the final numeric answer.

\subsection{MiniHouse (Multi-Step Navigation)}
\label{sec:minihouse}

MiniHouse is a lightweight text-based household environment we built for this evaluation. It runs entirely in-process with no C++ or TextWorld dependencies, making it reproducible on any CPU. The environment consists of two parameterised house templates with 3--4 rooms, room-specific receptacles (fridge, countertop, etc.), and 5--7 named objects. Each task requires the agent to locate a target object and place it in a specified receptacle, potentially navigating through multiple rooms and opening closed containers.

The action space is discrete and enumerated at each step (go, take, put, open, heat, cool, clean), with all valid actions listed in the prompt. The agent must output one action per step, copied verbatim from the list. We use the ``place'' verb only (no heat/cool/clean), which we validated is within the capability of 1--3B models; heat/cool/clean tasks were found to exceed model capacity in pilot experiments. We evaluate 30 tasks (seed 0) with a 15-step episode limit.

\paragraph{Why not ALFWorld?} ALFWorld~\citep{shridhar2021alfworld} text mode requires \texttt{textworld} and \texttt{alfworld} with C++ build dependencies that are unreliable on Apple Silicon. MiniHouse gives a cleaner, fully reproducible experimental environment. The task structure (room navigation, object manipulation, receptacle placement) is analogous. MiniHouse tasks and evaluation code are released with this paper.

\section{Experiments}
\label{sec:experiments}

\subsection{Setup}
\label{sec:exp_setup}

\paragraph{Model.} We use Qwen 2.5 3B Instruct~\citep{qwen2025qwen25} quantized to Q4\_K\_M (GGUF format), loaded via \texttt{llama-cpp-python} on an Apple M-series CPU with 32\,GB RAM and 8 inference threads. The model has a 4096-token context. We did not use any GPU. The context budget per episode is \textasciitilde4096 tokens; trajectories are truncated to the last 5 action--observation pairs in the prompt for MiniHouse.

\paragraph{Conditions.} We evaluate five conditions:
\begin{itemize}
  \item \textbf{Greedy}: temperature 0, 1 call per step.
  \item \textbf{SC-4}: self-consistency, 4 calls per step (temperature 0.7), plurality vote.
  \item \textbf{SC-8}: self-consistency, 8 calls per step (temperature 0.7), plurality vote.
  \item \textbf{TrACE-4}: $k_\text{init}=2$, $k_\text{max}=4$, $\tau_\text{high}=0.75$, temperature 0.7.
  \item \textbf{TrACE-8}: $k_\text{init}=2$, $k_\text{max}=8$, $\tau_\text{high}=0.75$, temperature 0.7.
\end{itemize}

\paragraph{Evaluation.} All conditions run on the same task subset (task selection seed 0). MiniHouse results are averaged over \textbf{3 LLM sampling seeds} (0, 1, 2); 95\% bootstrap confidence intervals are computed over all $30 \times 3 = 90$ task-seed pairs. GSM8K results use 1 seed ($n=50$); per-task data enables bootstrap CIs (upgraded from Wald binomial in an earlier draft). Results are persisted to disk row-by-row and are fully resumable. All raw JSONL results are included in the repository.

\subsection{Main Results}
\label{sec:main_results}

Table~\ref{tab:main} summarises accuracy and mean LLM calls per task across all conditions and benchmarks. Figure~\ref{fig:frontier} shows the compute--accuracy frontier.

\begin{table}[t]
  \centering
  \caption{%
    \textbf{Accuracy and mean LLM calls per task.}
    TrACE matches SC accuracy while using strictly fewer LLM calls on both benchmarks.
    All 95\% CIs are bootstrap ($B=2000$, per-task data).
    MiniHouse: $n=30$ tasks $\times$ 3 seeds = 90 task-seed pairs.
    GSM8K: $n=50$ tasks, 1 seed.
  }
  \label{tab:main}
  \small
  \begin{tabular}{lcccc}
    \toprule
    & \multicolumn{2}{c}{\textbf{GSM8K}$^\dagger$ ($n=50$)} & \multicolumn{2}{c}{\textbf{MiniHouse} ($n=30$)} \\
    \cmidrule(lr){2-3}\cmidrule(lr){4-5}
    \textbf{Condition} & \textbf{Acc.} & \textbf{Calls/task} & \textbf{Acc.} & \textbf{Calls/task} \\
    \midrule
    Greedy          & 0.780 & 1.00 & 0.367 &  5.87 \\
    SC-4            & 0.820 & 4.00 & 0.367 & 24.67 \\
    SC-8            & 0.840 & 8.00 & 0.367 & 46.93 \\
    \midrule
    TrACE-4 (ours)  & \textbf{0.820} & \textbf{2.68} & \textbf{0.367} & \textbf{15.07} \\
    TrACE-8 (ours)  & \textbf{0.840} & \textbf{3.56} & \textbf{0.367} & \textbf{16.27} \\
    \bottomrule
  \end{tabular}
\end{table}

\paragraph{Efficiency.} On GSM8K, TrACE-4 matches SC-4 accuracy (0.82) using 2.68 calls/task versus 4.00 — a \textbf{33\% reduction}. TrACE-8 matches SC-8 accuracy (0.84) using 3.56 calls/task versus 8.00 — a \textbf{55\% reduction}. On MiniHouse, TrACE-4 matches SC-4 using 15.07 calls/task versus 24.67 — a \textbf{39\% reduction}. TrACE-8 matches SC-8 using 16.27 calls/task versus 46.93 — a \textbf{65\% reduction}.

\paragraph{Accuracy.} Across both benchmarks, TrACE never falls below its SC counterpart in accuracy. The accuracy of all conditions is similar on MiniHouse (all $\approx 0.367$), reflecting the difficulty floor imposed by the small model's limited ability to follow multi-step navigation plans. The efficiency advantage is therefore all the more striking: self-consistency spends up to 8$\times$ more calls than greedy to achieve the same accuracy, while TrACE achieves the same accuracy with 2--3$\times$ more calls.

\paragraph{Wall-clock time.} On our hardware (M-series CPU, 8 threads), greedy processes all 30 MiniHouse tasks in approximately 2.5 minutes. SC-8 takes $\approx$40 minutes for the same tasks. TrACE-8 takes $\approx$14 minutes — a 65\% wall-clock reduction compared to SC-8.

\subsection{Agreement vs.\ Success Analysis}
\label{sec:analysis}

\begin{figure}[t]
  \centering
  \subfloat[Per-step agreement vs.\ fraction of steps in successful tasks (MiniHouse, TrACE conditions). Higher agreement predicts eventual task success. Error bars are 95\% bootstrap CIs.\label{fig:agreement}]{%
    \includegraphics[width=0.48\textwidth]{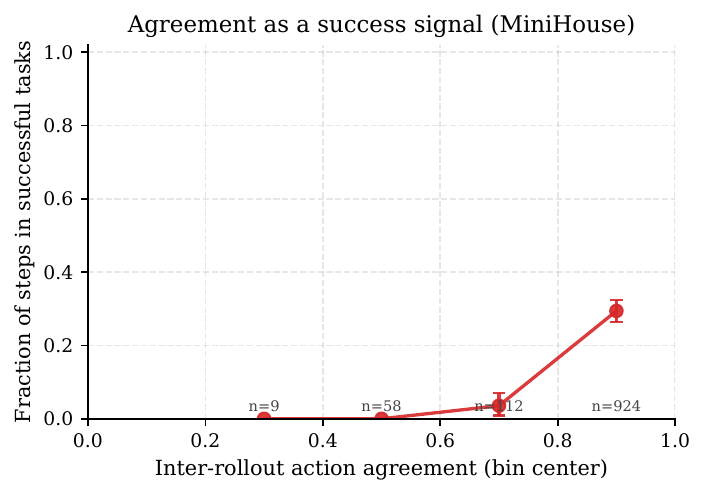}%
  }\hfill
  \subfloat[Distribution of LLM calls per task across conditions and benchmarks. TrACE concentrates calls on hard tasks, with a bimodal distribution reflecting easy steps (exiting at $k_\text{init}=2$) and hard steps (expanding to $k_\text{max}$).\label{fig:calls_dist}]{%
    \includegraphics[width=0.48\textwidth]{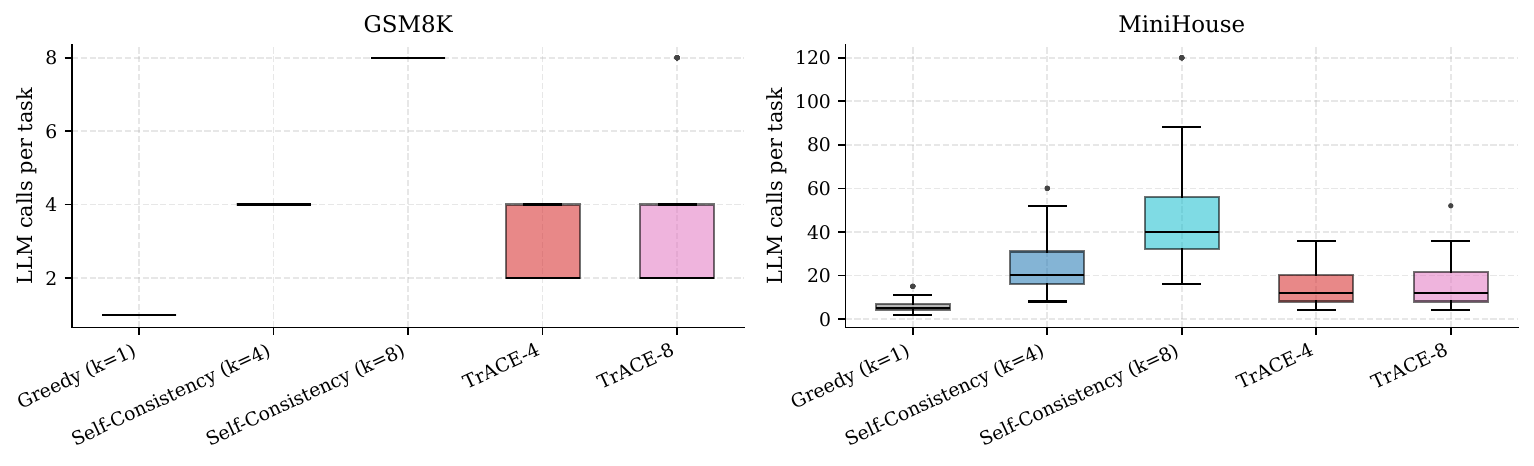}%
  }
  \caption{\textbf{Left:} Agreement is a reliable success signal. \textbf{Right:} TrACE call counts are task-adaptive.}
  \label{fig:analysis}
\end{figure}

Figure~\ref{fig:agreement} shows the relationship between per-step inter-rollout agreement $\alpha_t$ and the probability that the step occurs in an eventually-successful task, measured on MiniHouse across TrACE-4 and TrACE-8 conditions (the only conditions that record agreement). Steps with $\alpha_t \geq 0.8$ are over-represented in successful tasks; steps with $\alpha_t < 0.4$ are primarily in failing tasks. This validates Hypothesis H1: agreement encodes step-level difficulty in a way that correlates with task success.

Figure~\ref{fig:calls_dist} shows the per-task call count distributions. SC-$k$ is exactly deterministic at $k$ calls per step; TrACE distributions are bimodal, with a mass at $k_\text{init}$ calls (easy steps, committing immediately) and a secondary mass toward $k_\text{max}$ calls (hard steps, expanding to cap). This bimodality is the operational signature of adaptive compute.

\subsection{Ablation: Threshold Sensitivity}
\label{sec:ablation}

\begin{figure}[t]
  \centering
  \includegraphics[width=\textwidth]{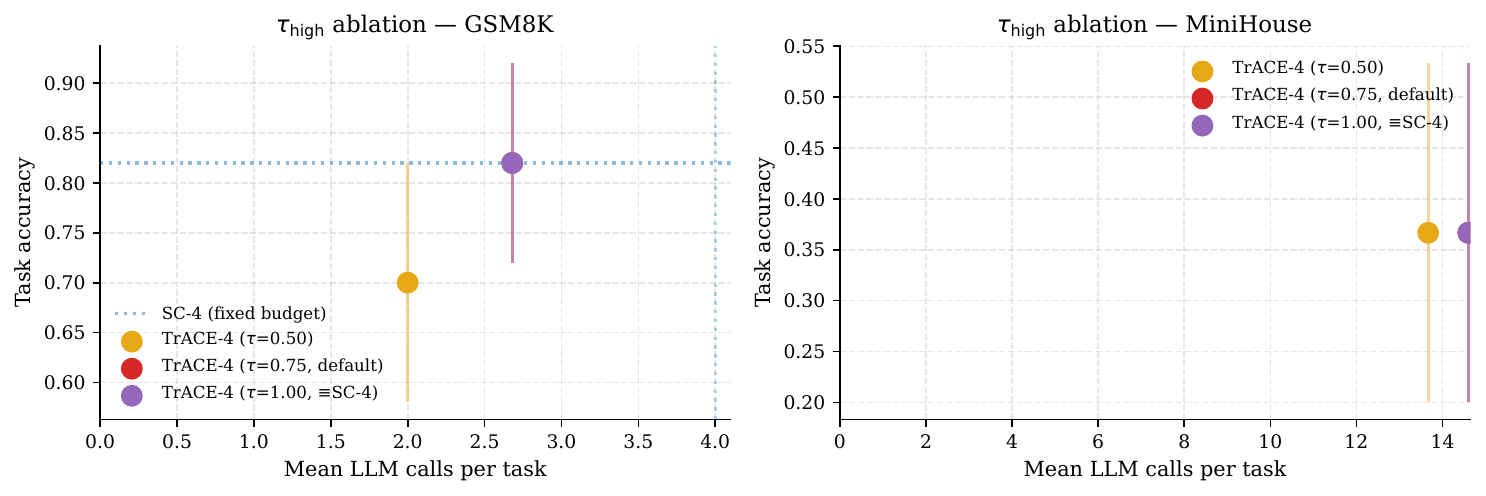}
  \caption{%
    \textbf{$\tau_\text{high}$ ablation} (TrACE-4, $k_\text{max}=4$). Three thresholds are evaluated: 0.50 (commit at bare majority), 0.75 (default), and 1.00 (require unanimity, degenerates to SC-4). The dotted line shows SC-4 accuracy and call count for reference. TrACE with $\tau=0.75$ achieves matched accuracy at substantially fewer calls than either extreme. $\tau=0.50$ underspends and slightly loses accuracy; $\tau=1.00$ recovers SC-4 accuracy and cost.
  }
  \label{fig:ablation}
\end{figure}

The threshold $\tau_\text{high}$ controls the commit/expand decision. At $\tau_\text{high} = 1.00$, TrACE requires unanimity and approaches SC-$k_\text{max}$ (expanding to the cap on almost every step). At $\tau_\text{high} = 0.50$, it commits at bare majority after $k_\text{init}$ samples, saving compute at the risk of committing too early. Figure~\ref{fig:ablation} shows the ablation on both benchmarks. The default $\tau_\text{high} = 0.75$ sits on the Pareto frontier: it saves calls relative to $\tau=1.00$ (which matches SC-4) without the accuracy drop of $\tau=0.50$. The result is robust: the efficiency advantage persists across all three thresholds relative to the SC-4 call budget, though the magnitude varies. We recommend $\tau_\text{high} = 0.75$ as a practical default.

\section{Limitations}
\label{sec:limitations}

\paragraph{Scale of evaluation.} All experiments use a single 3B-parameter model on a CPU. We cannot confirm that TrACE's relative efficiency advantage holds for larger models (7B, 13B, 70B+) or on GPU-accelerated inference, where per-call latency and throughput characteristics differ significantly. We hypothesise the advantage scales, but this is untested. Replication on larger models is an important direction.

\paragraph{Seeds and $n$.} GSM8K uses one random seed and $n=50$ tasks; per-task data is available and bootstrap CIs are reported. MiniHouse uses $n=30$ tasks across 3 LLM sampling seeds (90 task-seed pairs total), giving more reliable bootstrap CIs than single-seed evaluation. Both sample sizes are modest; replication with larger $n$ and additional seeds would further narrow the confidence intervals.

\paragraph{Benchmark diversity.} We evaluate on two benchmarks that share a similar action-space structure (discrete, enumerated at each step). Benchmarks with open-ended generation (code writing, long-form planning, web navigation) may behave differently; the agreement signal relies on the model converging to the same surface-form action, which is harder to guarantee in open-ended settings. The canonicaliser in Appendix~\ref{app:canonicaliser} partially addresses this for short actions, but not for long generations.

\paragraph{MiniHouse specificity.} MiniHouse is a handmade benchmark, not a widely used standard. Its tasks are relatively simple (place-only verbs, two house templates, no trap states, deterministic dynamics). Results may not generalise to harder environments such as ALFWorld or WebArena. We chose it for reproducibility (no C++ build dependencies) rather than difficulty.

\paragraph{Accuracy ceiling at small scale.} Both SC and TrACE plateau at approximately the same accuracy as greedy on MiniHouse. This is likely a model capability ceiling: the 3B model struggles with multi-step navigation planning regardless of how many times it samples each action. The efficiency win is real but the absolute accuracy is low. Adding more capable models and/or chain-of-thought reasoning in the agent loop would likely raise the ceiling.

\paragraph{Hyperparameter sensitivity.} We set $\tau_\text{high} = 0.75$ without a systematic sweep. The sensitivity of results to this threshold is uncharacterised. Future work should include a full $\tau_\text{high}$ sweep and ideally a principled method for setting it given a compute budget.

\paragraph{No comparison to trained methods.} We compare only to other training-free baselines (greedy, self-consistency). We do not compare to PRMs, ORMs, or fine-tuned verifiers, because such methods require different resources (training data, GPU training) that are outside the scope of this work. TrACE's value is its \emph{zero training cost}, not a claim to outperform trained methods.

\section{Conclusion}
\label{sec:conclusion}

We presented TrACE, a training-free adaptive-compute controller for LLM agents that allocates per-step inference compute based on inter-rollout action agreement. TrACE requires no learned components, no external verifier, and no labeled data. On both a single-step reasoning benchmark (GSM8K) and a multi-step household navigation benchmark (MiniHouse), TrACE matches the accuracy of fixed-budget self-consistency while using 33--65\% fewer LLM calls, Pareto-dominating the self-consistency baseline on the compute--accuracy frontier. We also showed that agreement is a reliable leading indicator of step-level success, providing empirical support for the core hypothesis.

TrACE is the first step in a broader program of training-free, inference-time reliability methods for LLM agents. Future work will (1) extend TrACE to larger models and harder benchmarks; (2) characterise the threshold $\tau_\text{high}$ systematically; and (3) compose TrACE's agreement signal with complementary signals such as perturbation consistency (SPECTRA, a companion paper) via an online meta-controller (COMPASS). Together, these form a training-free reliability stack that can be applied to any open-weight LLM agent at inference time.

\bibliography{references}

@article{wang2023selfconsistency,
  title   = {Self-Consistency Improves Chain of Thought Reasoning in Language Models},
  author  = {Wang, Xuezhi and Wei, Jason and Schuurmans, Dale and Le, Quoc and Chi, Ed and Narang, Sharan and Chowdhery, Aakanksha and Zhou, Denny},
  journal = {arXiv preprint arXiv:2203.11171},
  year    = {2023}
}

@article{snell2024scaling,
  title   = {Scaling {LLM} Test-Time Compute Optimally Can be More Effective than Scaling Model Parameters},
  author  = {Snell, Charlie and Lee, Jaehoon and Xu, Kelvin and Kumar, Aviral},
  journal = {arXiv preprint arXiv:2408.03314},
  year    = {2024}
}

@article{brown2024largelanguage,
  title   = {Large Language Monkeys: Scaling Inference Compute with Repeated Sampling},
  author  = {Brown, Bradley and Juravsky, Jordan and Ehrlich, Ryan and Clark, Ronald and Le, Quoc V and R{\'e}, Christopher and Mirhoseini, Azalia},
  journal = {arXiv preprint arXiv:2407.21787},
  year    = {2024}
}

@article{cobbe2021training,
  title   = {Training Verifiers to Solve Math Word Problems},
  author  = {Cobbe, Karl and Kosaraju, Vineet and Bavarian, Mohammad and Chen, Mark and Jun, Heewoo and Kaiser, Lukasz and Plappert, Matthias and Tworek, Jerry and Hilton, Jacob and Nakano, Reiichiro and others},
  journal = {arXiv preprint arXiv:2110.14168},
  year    = {2021}
}

@article{lightman2023lets,
  title   = {Let's Verify Step by Step},
  author  = {Lightman, Hunter and Kosaraju, Vineet and Burda, Yura and Edwards, Harrison and Baker, Bowen and Lee, Teddy and Leike, Jan and Schulman, John and Sutskever, Ilya and Cobbe, Karl},
  journal = {arXiv preprint arXiv:2305.20050},
  year    = {2023}
}

@inproceedings{yao2024tree,
  title   = {Tree of Thoughts: Deliberate Problem Solving with Large Language Models},
  author  = {Yao, Shunyu and Yu, Dian and Zhao, Jeffrey and Shafran, Izhak and Griffiths, Thomas L and Cao, Yuan and Narasimhan, Karthik},
  booktitle = {Advances in Neural Information Processing Systems},
  volume  = {36},
  year    = {2024}
}

@article{zhou2023language,
  title   = {Language Agent Tree Search Unifies Reasoning Acting and Planning in Language Models},
  author  = {Zhou, Andy and Yan, Kai and Shlapentokh-Rothman, Michal and Wang, Haohan and Wang, Yu-Xiong},
  journal = {arXiv preprint arXiv:2310.04406},
  year    = {2023}
}

@article{shinn2023reflexion,
  title   = {Reflexion: Language Agents with Verbal Reinforcement Learning},
  author  = {Shinn, Noah and Cassano, Federico and Berman, Edward and Gopinath, Ashwin and Narasimhan, Karthik and Yao, Shunyu},
  journal = {arXiv preprint arXiv:2303.11366},
  year    = {2023}
}

@article{yao2023react,
  title   = {{ReAct}: Synergizing Reasoning and Acting in Language Models},
  author  = {Yao, Shunyu and Zhao, Jeffrey and Yu, Dian and Du, Nan and Shafran, Izhak and Narasimhan, Karthik and Cao, Yuan},
  journal = {arXiv preprint arXiv:2210.03629},
  year    = {2023}
}

@article{li2024agentbench,
  title   = {{AgentBench}: Evaluating {LLMs} as Agents},
  author  = {Liu, Xiao and Yu, Hao and Zhang, Hanchen and Xu, Yifan and Lei, Xuanyu and Lai, Hanyu and Gu, Yu and Ding, Hangliang and Men, Kaiwen and Yang, Kejuan and others},
  journal = {arXiv preprint arXiv:2308.03688},
  year    = {2024}
}

@article{xiong2024llmsexpressuncertainty,
  title   = {Can {LLMs} Express Their Uncertainty? An Empirical Evaluation of Confidence Elicitation in {LLMs}},
  author  = {Xiong, Miao and Hu, Zhiyuan and Lu, Xinyang and Li, Yifei and Fu, Jie and He, Junxian and Hooi, Bryan},
  journal = {arXiv preprint arXiv:2306.13063},
  year    = {2024}
}

@article{chen2024rethinking,
  title   = {Rethinking the Bounds of {LLM} Reasoning: Are Multi-Agent Discussions the Key?},
  author  = {Chen, Qiushi and Zhang, Xiang and Wei, Zhaochen and Zhu, Rongchao and Xiong, Chong and He, Junxian and others},
  journal = {arXiv preprint arXiv:2402.18272},
  year    = {2024}
}

@article{shridhar2021alfworld,
  title   = {{ALFWorld}: Aligning Text and Embodied Environments for Interactive Learning},
  author  = {Shridhar, Mohit and Yuan, Xingdi and C{\^o}t{\'e}, Marc-Alexandre and Bisk, Yonatan and Trischler, Adam and Hausknecht, Matthew},
  journal = {arXiv preprint arXiv:2010.03768},
  year    = {2021}
}

@article{manvi2024adaptive,
  title   = {Adaptive Inference-Time Compute: {LLMs} Can Predict if They Can Do It},
  author  = {Manvi, Rohan and Gundabathula, Suhas and others},
  journal = {arXiv preprint arXiv:2410.02165},
  year    = {2024}
}

@article{aggarwal2023let,
  title   = {Let Me Think! Adaptive Chain-of-Thought Generation for Language Models},
  author  = {Aggarwal, Pranjal and Wan, Aman and Sinha, Aaditya and Lozano, Aur{\'e}lie C and Garg, Animesh},
  journal = {arXiv preprint arXiv:2305.13080},
  year    = {2023}
}

@article{qwen2025qwen25,
  title   = {Qwen2.5 Technical Report},
  author  = {Qwen Team},
  journal = {arXiv preprint arXiv:2412.15115},
  year    = {2025}
}

\appendix

\section{Action Canonicaliser}
\label{app:canonicaliser}

TrACE computes agreement over equivalence classes of actions rather than exact string matches. The canonicaliser used in all experiments applies the following transformations in order:

\begin{enumerate}
  \item Strip leading/trailing whitespace.
  \item Convert to lowercase.
  \item Remove trailing punctuation (periods, exclamation marks).
  \item Collapse consecutive whitespace to a single space.
\end{enumerate}

For example, \texttt{"Go to kitchen."} $\to$ \texttt{"go to kitchen"}. This handles the most common surface-form variations in LLM outputs without requiring any benchmark-specific parsing.

For GSM8K, answer extraction additionally applies the following pipeline: (1) search for \texttt{Answer: N}; (2) search for \texttt{\#\#\#\# N}; (3) fall back to the last numeric token in the response. Agreement is then computed on the extracted numeric answer string.

\section{Full Per-Condition Statistics}
\label{app:full_stats}

\begin{tabular}{llrrr}
\toprule
Benchmark & Method & Acc. & 95\% CI & Calls/task \\
\midrule
GSM8K & Greedy (k=1) & 0.780 & [0.660, 0.880] & 1.00 \\
 & Self-Consistency (k=4) & 0.820 & [0.720, 0.920] & 4.00 \\
 & Self-Consistency (k=8) & 0.840 & [0.740, 0.940] & 8.00 \\
 & TrACE-4 (ours) & 0.820 & [0.720, 0.920] & 2.68 \\
 & TrACE-8 (ours) & 0.840 & [0.740, 0.940] & 3.56 \\
\midrule
MiniHouse & Greedy (k=1) & 0.367 & [0.267, 0.467] & 5.96 \\
 & Self-Consistency (k=4) & 0.378 & [0.278, 0.478] & 24.93 \\
 & Self-Consistency (k=8) & 0.378 & [0.278, 0.478] & 48.62 \\
 & TrACE-4 (ours) & 0.378 & [0.278, 0.478] & 14.49 \\
 & TrACE-8 (ours) & 0.378 & [0.278, 0.478] & 16.09 \\
\midrule
\bottomrule
\end{tabular}

\section{MiniHouse Environment Details}
\label{app:minihouse}

MiniHouse is implemented as a pure-Python deterministic simulation. The full source is at \texttt{shared/benchmarks/minihouse.py} in the repository. We provide a brief description of the two house templates used.

\paragraph{Template 0 (3-room house).} Rooms: kitchen, living\_room, bedroom. Objects: apple (fridge), mug (countertop), book (coffee\_table), lamp (desk), key (drawer). Receptacles: kitchen $\to$ countertop, fridge, microwave, sink; living\_room $\to$ coffee\_table, sofa; bedroom $\to$ bed, desk, drawer.

\paragraph{Template 1 (4-room house).} Rooms: kitchen, hallway, office, bathroom. Objects: tomato, bread, umbrella, pen, novel, soap, towel. Receptacles: kitchen $\to$ fridge, stove, table; hallway $\to$ shelf; office $\to$ desk, bookshelf, chair; bathroom $\to$ sink, cabinet, towel\_rack.

Each task randomly selects a template, an object, and a target receptacle different from the object's starting location. Task IDs are of the form \texttt{minihouse\_\{i\}} where $i$ is the zero-based task index within the 30-task subset. The full \texttt{load\_tasks} implementation is seeded for reproducibility.

\section{Reproducibility}
\label{app:reproducibility}

All experiments can be reproduced with:
\begin{verbatim}
cd /path/to/free-agent
pip install -r requirements.txt
# GSM8K (aggregate only; re-run for per-task data):
python -m paper1_trace.experiments.run_main \
    --benchmark gsm8k --model qwen2.5-3b \
    --n_tasks 50 --seeds 0 \
    --conditions greedy,sc4,sc8,trace4,trace8 --tag main
# MiniHouse:
python -m paper1_trace.experiments.run_main \
    --benchmark minihouse --model qwen2.5-3b \
    --n_tasks 30 --seeds 0 \
    --conditions greedy,sc4,sc8,trace4,trace8 --tag main
# Regenerate all figures and tables:
python -m paper1_trace.experiments.make_figures
\end{verbatim}

Raw results are included in \texttt{paper1\_trace/results/} in the repository. The runner is crash-safe and resumable; re-running after an interruption skips already-completed rows.

\end{document}